\pdfoutput=1 
\documentclass[11pt,a4paper]{article}

\usepackage[hyperref]{naaclhlt2019}
\usepackage{times}
\usepackage{latexsym}

\usepackage{url}

\usepackage{graphicx}
\usepackage{textcomp,    
            xspace}
\newcommand\la{\textlangle\xspace}  
\newcommand\ra{\textrangle\xspace}
\usepackage{multirow}

\usepackage{array}  
\newcolumntype{L}[1]{>{\raggedright\let\newline\\\arraybackslash\hspace{0pt}}p{#1}}
\newcolumntype{R}[1]{>{\raggedleft\let\newline\\\arraybackslash\hspace{0pt}}p{#1}}

\aclfinalcopy 
\def\aclpaperid{***} 

\newcommand\blfootnote[1]{%
  \begingroup
  \renewcommand\thefootnote{}\footnote{#1}%
  \addtocounter{footnote}{-1}%
  \endgroup
}

\title{The Teacher-Student Chatroom Corpus}

\author{
  Andrew Caines\textsuperscript{1} \hspace{3mm}
  Helen Yannakoudakis\textsuperscript{2} \hspace{3mm}
  Helena Edmondson\textsuperscript{3} \hspace{3mm}
  Helen Allen\textsuperscript{4} \hspace{3mm} \\
  \bf 
  Pascual P\'{e}rez-Paredes\textsuperscript{5} \hspace{6mm}
  Bill Byrne\textsuperscript{6} \hspace{6mm}
  Paula Buttery\textsuperscript{1} \\
  \vspace{-2mm} \\
  \textsuperscript{1} ALTA Institute \& Computer Laboratory, University of Cambridge, U.K. \\ \texttt{\{andrew.caines|paula.buttery\}@cl.cam.ac.uk} \\
  \textsuperscript{2} Department of Informatics, King's College London, U.K. \\ \texttt{helen.yannakoudakis@kcl.ac.uk} \\
  \textsuperscript{3} Theoretical \& Applied Linguistics, University of Cambridge, U.K. \\ \texttt{hle24@cantab.ac.uk} \\
  \textsuperscript{4} Cambridge Assessment, University of Cambridge, U.K. \\ \texttt{allen.h@cambridgeenglish.org} \\
  \textsuperscript{5} Faculty of Education, University of Cambridge, U.K. \\ \texttt{pfp23@cam.ac.uk} \\
  \textsuperscript{6} Department of Engineering, University of Cambridge, U.K. \\ \texttt{bill.byrne@eng.cam.ac.uk}
}

\date{}

\begin{document}
\maketitle
\begin{abstract}
The Teacher-Student Chatroom Corpus (TSCC) is a collection of written conversations captured during one-to-one lessons between teachers and learners of English. The lessons took place in an online chatroom and therefore involve more interactive, immediate and informal language than might be found in asynchronous exchanges such as email correspondence. The fact that the lessons were one-to-one means that the teacher was able to focus exclusively on the linguistic abilities and errors of the student, and to offer personalised exercises, scaffolding and correction. The TSCC contains more than one hundred lessons between two teachers and eight students, amounting to 13.5K conversational turns and 133K words: it is freely available for research use. We describe the corpus design, data collection procedure and annotations added to the text. We perform some preliminary descriptive analyses of the data and consider possible uses of the TSCC.
\end{abstract}

\blfootnote{ This work is licensed under a Creative Commons Attribution-NonCommercial-ShareAlike 4.0 International Licence. Licence details: \url{http://creativecommons.org/licenses/by-nc-sa/4.0} }

\section{Introduction \& Related Work}

We present a new corpus of written conversations from one-to-one, online lessons between English language teachers and learners of English. This is the Teacher-Student Chat Corpus (TSCC) and it is openly available for research use\footnote{Available for download from \url{https://forms.gle/oW5fwTTZfZcTkp8v9}}. TSCC currently contains 102 lessons between 2 teachers and 8 students, which in total amounts to 13.5K conversational turns and 133K word tokens, and it will continue to grow if funding allows. 

The corpus has been annotated with grammatical error corrections, as well as discourse and teaching-focused labels, and we describe some early insights gained from analysing the lesson transcriptions. We also envisage future use of the corpus to develop dialogue systems for language learning, and to gain a deeper understanding of the teaching and learning process in the acquisition of English as a second language.

We are not aware of any such existing corpus, hence we were motivated to collect one. To the best of our knowledge, the TSCC is the first to feature one-to-one online chatroom conversations between teachers and students in an English language learning context. There are of course many conversation corpora prepared with both close discourse analysis and machine learning in mind. For instance, the Cambridge and Nottingham Corpus of Discourse in English (CANCODE) contains spontaneous conversations recorded in a wide variety of informal settings and has been used to study the grammar of spoken interaction \cite{carter-mccarthy-1997}. Both versions 1 and 2 of the British National Corpus feature transcriptions of spoken conversation captured in settings ranging from parliamentary debates to casual discussion among friends and family \cite{bnc,love-et-al-2017}.

Corpora based on educational interactions, such as lectures and small group discussion, include the widely-used Michigan Corpus of Academic Spoken English (MICASE) \cite{micase}, TOEFL 2000 Spoken and Written Academic Language corpus \cite{biber-et-al-2004}, and Limerick Belfast Corpus of Academic Spoken English (LI-BEL) \cite{okeeffe-walsh-2012}. Corpora like the ones listed so far, collected with demographic and linguistic information about the contributors, enable the study of sociolinguistic and discourse research questions such as the interplay between lexical bundles and discourse functions \cite{csomay-2012}, the interaction of roles and goal-driven behaviour in academic discourse \cite{evison-2013}, and knowledge development at different stages of higher education learning \cite{atwood-et-al-2010}.

On a larger scale, corpora such as the Multi-Domain Wizard-of-Oz datasets (MultiWOZ) contain thousands of goal-directed dialogue collected through crowdsourcing and intended for the training of automated dialogue systems \cite{multiwoz,multiwoz2}. Other work has involved the collation of pre-existing conversations on the web, for example from Twitter \cite{ritter-et-al-2010}, Reddit \cite{schrading-et-al-2015}, and movie scripts \cite{danescu-lee-2011}. Such datasets are useful for training dialogue systems to respond to written inputs \textendash~so-called `chatbots' \textendash~which in recent years have greatly improved in terms of presenting some kind of personality, empathy and world knowledge \cite{blender}, where previously there had been relatively little of all three. The improvement in chatbots has caught the attention of, and in turn has been driven by, the technology industry, for they have clear commercial applications in customer service scenarios such as helplines and booking systems. 

As well as personality, empathy and world knowledge, if chatbots could also assess the linguistic proficiency of a human interlocutor, give pedagogical feedback, select appropriate tasks and topics for discussion, maintain a long-term memory of student language development, \emph{and} begin and close a lesson on time, that would be a teaching chatbot of sorts. We know that the list above represents a very demanding set of technological challenges, but the first step towards these more ambitious goals is to collect a dataset which allows us to analyse how language teachers operate, how they respond to student needs and structure a lesson. This dataset may indicate how we can begin to address the challenge of implementing a language teaching chatbot.

We therefore set out to collect a corpus of one-to-one teacher-student language lessons in English, since we are unaware of existing corpora of this type. The most similar corpora we know of are the Why2Atlas Human-Human Typed Tutoring Corpus which centred on physics tutoring \cite{rose-et-al-2003}, the chats collected between native speakers and learners of Japanese using a virtual reality university campus \cite{toyoda-harrison-2002}, and an instant messaging corpus between native speakers and learners of German \cite{hohn-2017}. This last corpus was used to design an other-initiated self-repair module in a dialogue system for language learning, the kind of approach which we aim to emulate in this project. In addition there is the CIMA dataset released this year which involves one-to-one written conversation between crowdworkers role-playing as teachers and students \cite{stasaski-et-al-2020}, but the fact that they are not genuinely teachers and learners of English taking part in real language lessons means that the data lack authenticity (albeit the corpus is well structured and useful for chatbot development).

\section{Corpus design}

We set out a design for the TSCC which was intended to be convenient for participants, efficient for data processing, and would allow us to make the data public. The corpus was to be teacher-centric: we wanted to discover how teachers deliver an English language lesson, adapt to the individual student, and offer teaching feedback to help students improve. On the other hand, we wanted as much diversity in the student group as possible, and therefore aimed to retain teachers during the data collection process as far as possible, but to open up student recruitment as widely as possible.

In order to host the lessons, we considered several well-known existing platforms, including Facebook Messenger, WhatsApp, and Telegram, but decided against these due firstly to concerns about connecting people unknown to each other, where the effect of connecting them could be long-lasting and unwanted (\emph{i.e.}\ ongoing messaging or social networking beyond the scope of the TSCC project). Secondly we had concerns that since those platforms retain user data to greater or lesser extent, we were requiring that study participants give up some personal information to third party tech firms \textendash~which they may already be doing, but we didn't want to \emph{require} this of the participants.

\begin{figure*}[t]
\centering
\includegraphics[width=\textwidth]{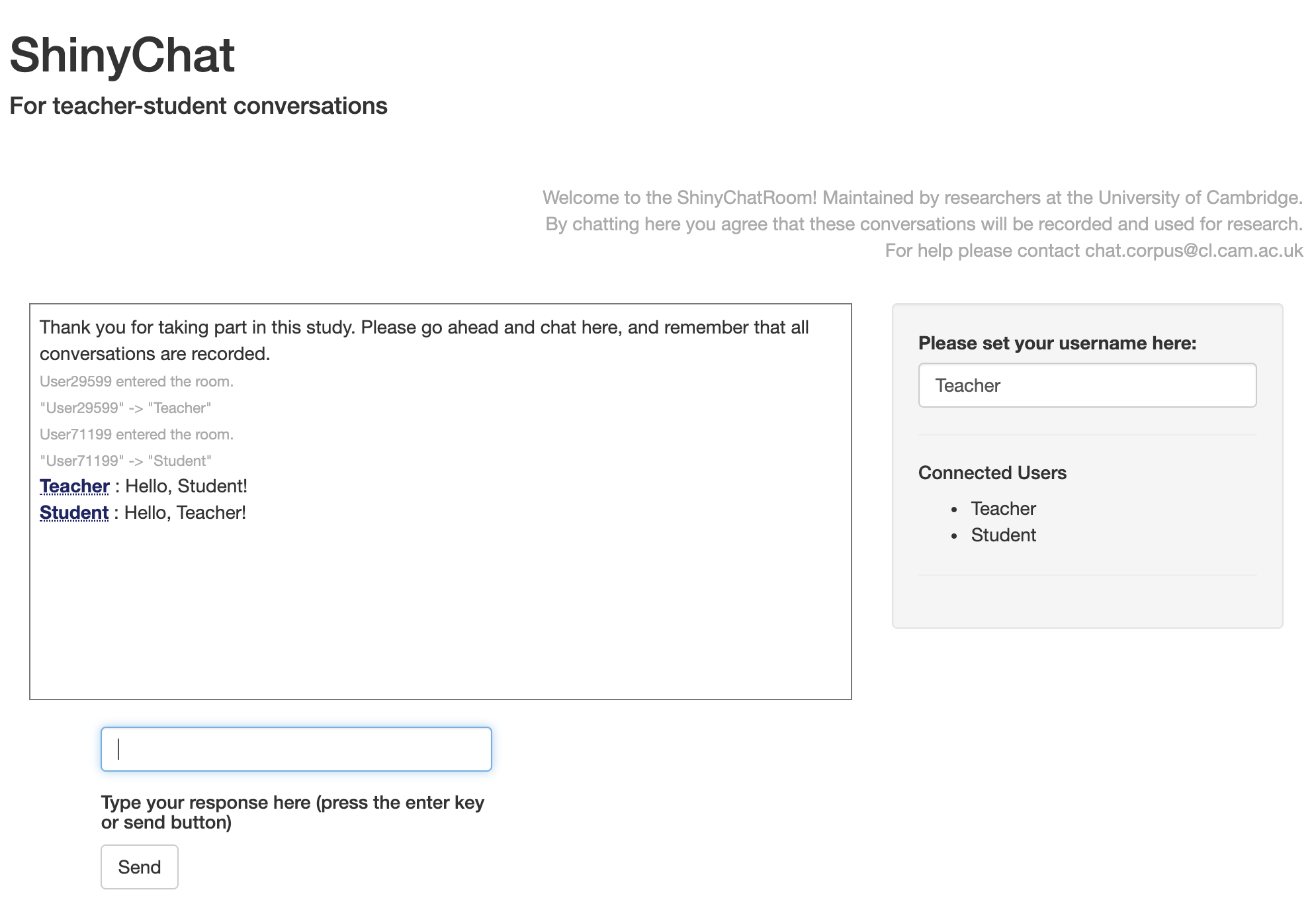}
\caption{Screenshot of the `ShinyChat' chatroom}
\label{fig:chatroom}
\end{figure*}

We consequently decided to use ephemeral chatrooms to host the lessons, and looked into using existing platforms such as Chatzy, but again had privacy concerns about the platform provider retaining their own copy of the lesson transcriptions (a stated clause in their terms and conditions) for unknown purposes. Thus we were led to developing our own chatroom in Shiny for R \cite{shiny}. In designing the chatroom we kept it as minimal and uncluttered as possible; it had little extra functionality but did the basics of text entry, username changes, and link highlighting.

Before recruiting participants, we obtained ethics approval from our institutional review board, on the understanding that lesson transcripts would be anonymised before public release, that participant information forms would be at an appropriate linguistic level for intermediate learners of English, and that there would be a clear procedure for participants to request deletion of their data if they wished to withdraw from the study. Funding was obtained in order to pay teachers for their participation in the study, whereas students were not paid for participation on the grounds that they were receiving a free one-to-one lesson.

\section{Data collection}

We recruited two experienced, qualified English language teachers to deliver the online lessons one hour at a time, on a one-to-one basis with students. The teacher-student pair were given access to the chatroom web application (Figure~\ref{fig:chatroom}) and we obtained a transcription of the lesson at the end of the lesson.

When signing up to take part in the study, all participants were informed that the contents of the lesson would be made available to researchers in an anonymised way, but to avoid divulging personally identifying information, or other information they did not wish to be made public. A reminder to this effect was displayed at the start of every chatroom lesson. As an extra precaution, we made anonymisation one part of the transcription annotation procedure; see the next section for further detail.

Eight students have so far been recruited to take part in the chatroom English lessons which form this corpus. All students participated in at least 2 lessons each (max=32, mean=12). Therefore one possible use of the corpus is to study longitudinal pedagogical effects and development of second language proficiency in written English chat. At the time of data collection, the students were aged 12 to 40, with a mean of 23 years. Their first languages are Japanese (2), Ukrainian (2), Italian, Mandarin Chinese, Spanish, and Thai.

We considered being prescriptive about the format of the one-hour lessons, but in the end decided to allow the teachers to use their teaching experience and expertise to guide the content and planning of lessons. This was an extra way of discovering how teachers structure lessons and respond to individual needs, while also observing what additional resources they call on (other websites, images, source texts, etc). When signing up to participate in the study, the students were able to express their preferences for topics and skills to focus on, information which was passed on to the teachers in order that they could prepare lesson content accordingly. Since most students return for several lessons with the teachers, we can also observe how the teachers guide them through the unwritten `curriculum' of learning English, and how students respond to this long-term treatment.

\section{Annotation}

The 102 collected lesson transcriptions have been annotated by an experienced teacher and examiner of English. The transcriptions were presented as spreadsheets, with each turn of the conversation as a new row, and columns for annotation values. There were several steps to the annotation process, listed and described below.

\vspace{1mm}
\noindent
\textbf{Anonymisation}: As a first step before any further annotation was performed, we replaced personal names with \la{TEACHER}\ra or \la{STUDENT}\ra placeholders as appropriate to protect the privacy of teacher and student participants. For the same reason we replaced other sensitive data such as a date-of-birth, address, telephone number or email address with \la{DOB}\ra, \la{ADDRESS}\ra, \la{TELEPHONE}\ra, \la{EMAIL}\ra. Finally, any personally identifying information \textendash~the mention of a place of work or study, description of a regular pattern of behaviour, \emph{etc} \textendash~was removed if necessary.

\vspace{1mm}
\noindent
\textbf{Grammatical error correction}: As well as the original turns of each participant, we also provide grammatically corrected versions of the student turns. The teachers make errors too, which is interesting in itself, but the focus of teaching is on the students and therefore we economise effort by correcting student turns only. The process includes grammatical errors, typos, and improvements to lexical choice. This was done in a minimal fashion to stay as close to the original meaning as possible. In addition, there can often be many possible corrections for any one grammatical error, a known problem in corpus annotation and NLP work on grammatical errors \cite{bryant-ng-2015}. The usual solution is to collect multiple annotations, which we have not yet done, but plan to. In the meantime, the error annotation is useful for grammatical error \emph{detection} even if \emph{correction} might be improved by more annotation.

\vspace{1mm}
\noindent
\textbf{Responding to}: This step involves the disentangling of conversational turns so that it was clear which preceding turn was being addressed, if it was not the previous one. As will be familiar from messaging scenarios, people can have conversations in non-linear ways, sometimes referring back to a turn long before the present one. For example, the teacher might write something in turn number 1, then something else in turn 2. In turn 3 the student responds to turn 2 \textendash~the previous one, and therefore an unmarked occurrence \textendash~but in turn 4 they respond to turn 1. The conversation `adjacency pairs' are thus non-linear, being 1\&4, 2\&3.

\vspace{1mm}
\noindent
\textbf{Sequence type}: We indicate major and minor shifts in conversational sequences \textendash~sections of interaction with a particular purpose, even if that purpose is from time-to-time more social than it is educational. Borrowing key concepts from the \textsc{conversation analysis} (CA) approach \cite{sacks-et-al-1974}, we seek out groups of turns which together represent the building blocks of the chat transcript: teaching actions which build the structure of the lessons.
\vspace{-2mm}
\begin{quote}
CA practitioners aim `to discover how participants understand and respond to one another in their turns at talk, with a central focus on how \emph{sequences of action} are generated' (\newcite{seedhouse-2004} quoting \newcite{hutchby-wooffitt-1988}, emphasis added).
\end{quote}
We define a number of sequence types listed and described below, firstly the major and then the minor types, or `sub-sequences':
\begin{itemize}
\setlength\itemsep{0em}
    \item Opening -- greetings at the start of a conversation; may also be found mid-transcript, if for example the conversation was interrupted and conversation needs to recommence.
    \item Topic \_\_\_ -- relates to the topic of conversation (minor labels complete this sequence type).
    \item Exercise -- signalling the start of a constrained language exercise (\emph{e.g.}\ `please look at textbook page 50', `let's look at the graph', \emph{etc}); can be controlled or freer practice (\emph{e.g.}\ gap-filling versus prompted re-use).
    \item Redirection -- managing the conversation flow to switch from one topic or task to another.
    \item Disruption -- interruption to the flow of conversation for some reason; for example because of loss of internet connectivity, telephone call, a cat stepping across the keyboard, and so on...
    \item Homework -- the setting of homework for the next lesson, usually near the end of the present lesson.
    \item Closing -- appropriate linguistic exchange to signal the end of a conversation.
    \item[] Below we list our minor sequence types, which complement the major sequence types:
    \begin{itemize}
    \setlength\itemsep{0em}
    \item Topic opening -- starting a new topic: will usually be a new sequence.
        \item Topic development -- developing the current topic: will usually be a new sub-sequence.
        \item Topic closure -- a sub-sequence which brings the current topic to a close.
        \item Presentation -- (usually the teacher) presenting or explaining a linguistic skill or knowledge component.
        \item Eliciting -- (usually the teacher) continuing to seek out a particular response or realisation by the student.
        \item Scaffolding -- (usually the teacher) giving helpful support to the student.
        \item Enquiry -- asking for information about a specific skill or knowledge component.
        \item Repair -- correction of a previous linguistic sequence, usually in a previous turn, but could be within a turn; could be correction of self or other.
        \item Clarification -- making a previous turn clearer for the other person, as opposed to `repair' which involves correction of mistakes.
        \item Reference -- reference to an external source, for instance recommending a textbook or website as a useful resource.
        \item Recap -- (usually the teacher) summarising a take-home message from the preceding turns.
        \item Revision -- (usually the teacher) revisiting a topic or task from a previous lesson.
    \end{itemize}
\end{itemize}
Some of these sequence types are exemplified in Table~\ref{tbl:bigex}.

\begin{table*}[t]
\begin{center}
\begin{tabular}{ c|c |L{4.4cm} |L{4.4cm} |r|l }
\bf Turn & \bf Role & \bf Anonymised & \bf Corrected & \bf Resp.to & \bf Sequence \\
\hline
1 & T & Hi there \la{STUDENT}\ra, all OK? & Hi there \la{STUDENT}\ra, all OK? & & opening \\
\hline
2 & S & Hi \la{TEACHER}\ra, how are you? & Hi \la{TEACHER}\ra, how are you? & & \\
\hline
3 & S & I did the exercise this morning & I did \emph{some} exercise this morning & & \\
\hline
4 & S & I have done, I guess & I have done, I guess & & repair \\
\hline
5 & T &  did is fine especially if you’re focusing on the action itself &  did is fine especially if you’re focusing on the action itself & & scaffolding \\
\hline
6 & T &  tell me about your exercise if you like! &  tell me about your exercise if you like! & 3 & topic.dev \\
\end{tabular}
\end{center}
\caption{\label{tbl:bigex} Example of numbered, anonymised and annotated turns in the TSCC (where role T=teacher, S=student, and `resp.to' means `responding to'); the student is here chatting about physical exercise. }
\end{table*}

\vspace{1mm}
\noindent
\textbf{Teaching focus}: Here we note what type of knowledge is being targeted in the new conversation sequence or sub-sequence. These usually accompany the sequence types, Exercise, Presentation, Eliciting, Scaffolding, Enquiry, Repair and Revision.
\begin{itemize}
\setlength\itemsep{0em}
    \item Grammatical resource -- appropriate use of grammar.
    \item Lexical resource -- appropriate and varied use of vocabulary.
    \item Meaning -- what words and phrases mean (in specific contexts).
    \item Discourse management -- how to be coherent and cohesive, refer to given information and introduce new information appropriately, signal discourse shifts, disagreement, and so on.
    \item Register -- information about use of language which is appropriate for the setting, such as levels of formality, use of slang or profanity, or intercultural issues.
    \item Task achievement -- responding to the prompt in a manner which fully meets requirements.
    \item Interactive communication -- how to structure a conversation, take turns, acknowledge each other's contributions, and establish common ground.
    \item World knowledge -- issues which relate to external knowledge, which might be linguistic (\emph{e.g.}\ cultural or pragmatic subtleties) or not (they might simply be relevant to the current topic and task).
    \item Meta knowledge -- discussion about the type of knowledge required for learning and assessment; for instance, `there’s been a shift to focus on X in teaching in recent years'.
    \item Typo - orthographic issues such as spelling, grammar or punctuation mistake
    \item Content -- a repair sequence which involves a correction in meaning; for instance, Turn 1: Yes, that's fine. Turn 2: Oh wait, no, it's not correct.
    \item Exam practice -- specific drills to prepare for examination scenarios.
    \item Admin -- lesson management, such as `please check your email' or `see page 75'.
\end{itemize}

\vspace{1mm}
\noindent
\textbf{Use of resource}: At times the teacher refers the student to materials in support of learning. These can be the chat itself \textendash~where the teacher asks the student to review some previous turns in that same lesson \textendash~or a textbook page, online video, social media account, or other website.

\vspace{1mm}
\noindent
\textbf{Student assessment}: The annotator, a qualified and experienced examiner of the English language, assessed the proficiency level shown by the student in each lesson. Assessment was applied according to the Common European Framework of Reference for Languages (CEFR)\footnote{\url{https://www.cambridgeenglish.org/exams-and-tests/cefr}}, with levels from A1 (least advanced) to C2 (most advanced). We anticipated that students would get more out of the lessons if they were already at a fairly good level, and therefore aimed our recruitment of participants at the intermediate level and above (CEFR B1 upwards). Assessment was applied in a holistic way based on the student's turns in each lesson: evaluating use of language (grammar and vocabulary), coherence, discourse management and interaction.

\begin{table}[t!]
\centering
\begin{tabular}{l|rrr}
  \bf Section & \bf Lessons & \bf Conv.turns & \bf Words \\
  \hline
  Teachers & 102 & 7632 & 93,602 \\
  Students & 102 & 5920 & 39,293 \\
  \hline
  All & 102 & 13,552 & 132,895 \\
\end{tabular}
\caption{ Number of lessons, conversational turns and words in the TSCC contributed by teachers, students and all combined. }
\label{tbl:tscc}
\end{table}

\begin{table}[t!]
\centering
\begin{tabular}{l|rrr}
  \bf Section & \bf Lessons & \bf Conv.turns & \bf Words \\
  \hline
  B1 & 36 & 1788 & 11,898 \\
  B2 & 37 & 2394 & 11,331 \\
  C1 & 29 & 1738 & 16,064 \\
  \hline
  Students & 102 & 5920 & 39,293 \\
\end{tabular}
\caption{ Number of lessons, conversational turns and words in the TSCC grouped by CEFR level. }
\label{tbl:studentcefr}
\end{table}

\begin{table*}[h]
\begin{center}
\begin{tabular}{ll|r|r|r}
& & \bf FCE & \bf CrowdED & \bf TSCC \\
\hline
\multirow{3}{*}{\bf Edit type} & Missing & 21.0\% & 13.9\% & 18.2\% \\
& Replacement & 64.4\% & 47.9\% & 72.3\% \\
& Unnecessary & 11.5\% & 38.2\% & 9.5\% \\
\hline
\multirow{24}{*}{\bf Error type} & Adjective & 1.4\% & 0.8\% & 1.5\% \\
& Adjective:form & 0.3\% & 0.06\% & 0.1\% \\
& Adverb & 1.9\% & 1.5\% & 1.6\% \\
& Conjunction & 0.7\% & 1.3\% & 0.2\% \\
& Contraction & 0.3\% & 0.4\% & 0.1\% \\
& Determiner & 10.9\% & 4.0\% & 12.4\% \\
& Morphology & 1.9\% & 0.6\% & 2.4\% \\
& Noun & 4.6\% & 5.8\% & 9.0\% \\
& Noun:inflection & 0.5\% & 0.01\% & 0.1\% \\
& Noun:number & 3.3\% & 1.0\% & 2.1\% \\
& Noun:possessive & 0.5\% & 0.1\% & 0.03\% \\
& Orthography & 2.9\% & 3.0\% & 6.7\% \\
& Other & 13.3\% & 61.0\% & 28.4\% \\
& Particle & 0.3\% & 0.5\% & 0.6\% \\
& Preposition & 11.2\% & 2.9\% & 7.4\% \\
& Pronoun & 3.5\% & 1.2\% & 2.9\% \\
& Punctuation & 9.7\% & 8.7\% & 0.9\% \\
& Spelling & 9.6\% & 0.3\% & 6.0\% \\
& Verb & 7.0\% & 3.1\% & 6.7\% \\
& Verb:form & 3.6\% & 0.4\% & 2.9\% \\
& Verb:inflection & 0.2\% & 0.01\% & 0.1\% \\
& Verb:subj-verb-agr & 1.5\% & 0.3\% & 1.8\% \\
& Verb:tense & 6.0\% & 1.1\% & 4.8\% \\
& Word order & 1.8\% & 1.2\% & 1.0\% \\
\hline
\multirow{3}{*}{\bf Corpus stats} & Texts & 1244 & 1108 & 102 \\
& Words & 531,416 & 39,726 & 132,895 \\
& Total edits & 52,671 & 8454 & 3800 \\
\end{tabular}
\end{center}
\caption{\label{tbl:errors} The proportional distribution of error types determined by grammatical error correction of texts in the TSCC. Proportions supplied for the FCE Corpus for comparison, from \newcite{bryant-et-al-2019}, and a subset of the \textsc{CrowdED} Corpus (for a full description of error types see \newcite{bryant-et-al-2017}) }
\end{table*}

In Table~\ref{tbl:bigex} we exemplify many of the annotation steps described above with an excerpt from the corpus. We show several anonymised turns from one of the lessons, with turn numbers, participant role, error correction, `responding to' when not the immediately preceding turn, and sequence type labels. Other labels such as teaching focus and use of resource are in the files but not shown here. The example is not exactly how the corpus texts are formatted, but it serves to illustrate: the \textsc{readme} distributed with the corpus further explains the contents of each annotated chat file.

The annotation of the features described above may in the long-term enable improved dialogue systems for language learning, and for the moment we view them as a first small step towards that larger goal. We do not yet know which features will be most useful and relevant for training such dialogue systems, but that is the purpose of collecting wide-ranging annotation. The corpus size is still relatively small, and so for the time being they allow us to focus on the analysis of one-to-one chat lessons and understand how such lessons are structured by both teacher and student.

\begin{figure*}[h]
    \centering
    \includegraphics[width=\textwidth]{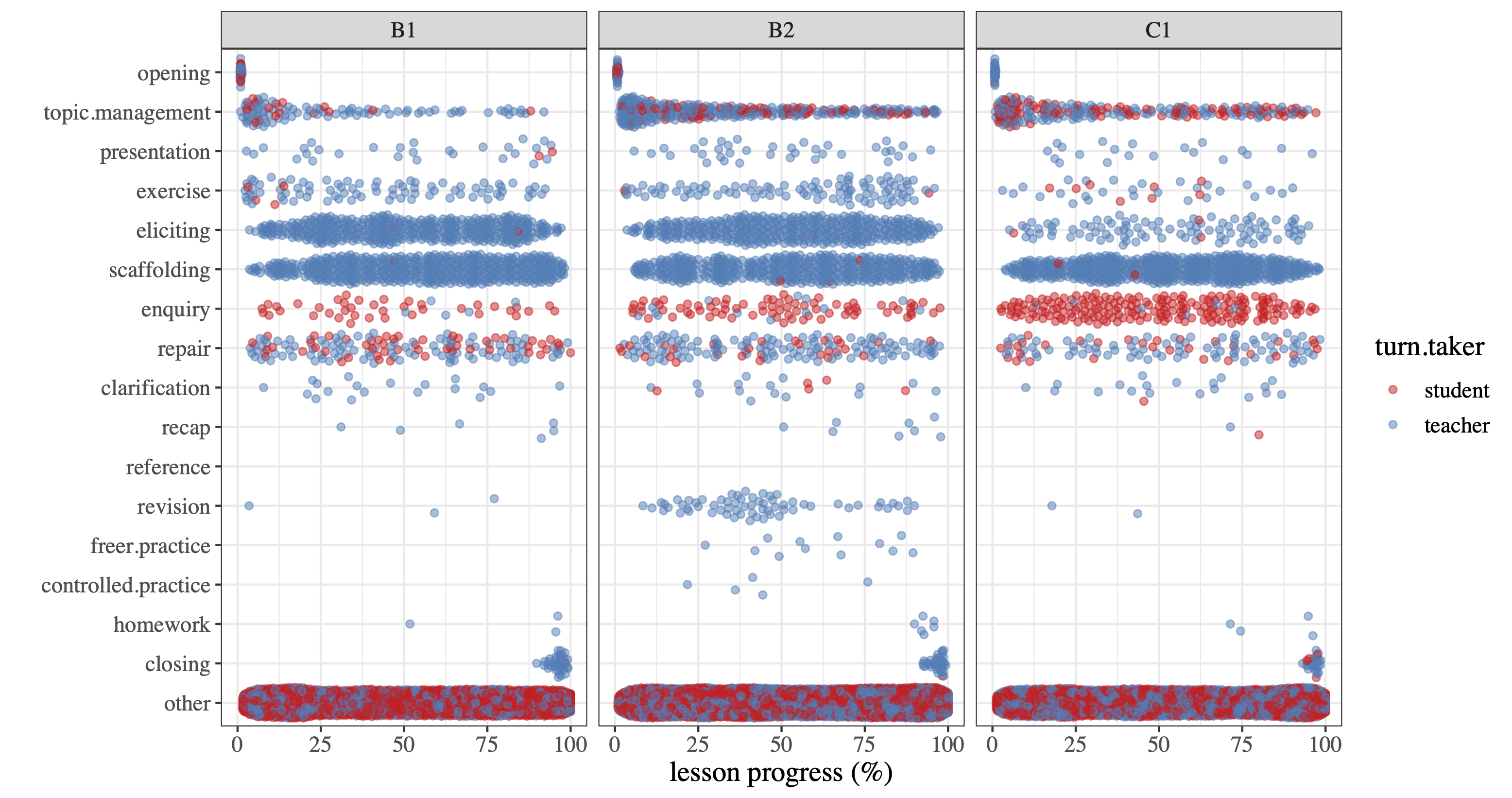}
    \caption{ Selected sequence types in the TSCC, one plot per CEFR level, teachers as blue points, students as red; types on the $y$-axis and lesson progress on the $x$-axis (\%). `Other' represents all non-sequence-starting turns in the corpus. }
    \label{fig:sequences}
\end{figure*}

\section{Corpus analysis}

In Table~\ref{tbl:tscc} we report the overall statistics for TSCC in terms of lessons, conversational turns, and number of words (counted as white-space delimited tokens). We also show these statistics for the teacher and student groups separately. It is unsurprising that the teachers contribute many more turns and words to the chats than their students, but perhaps surprising just how much more they contribute. Each lesson was approximately one hour long and amounted to an average of 1300 words.

In Table~\ref{tbl:studentcefr} we show these same statistics for the student group only, and this time subsetting the group by the CEFR levels found in the corpus: B1, B2 and C1. As expected, no students were deemed to be of CEFR level A1 or A2 in their written English, and the majority were of the intermediate B1 and B2 levels. It is notable that the B2 students in the corpus contribute many more turns than their B1 counterparts, but fewer words. The C1 students \textendash~the least numerous group \textendash~contribute the fewest turns of all groups but by far the most words. All the above might well be explained by individual variation and/or by teacher task and topic selection (\emph{e.g.}\ setting tasks which do or do not invite longer responses) per the notion of `opportunity of use' \textendash~what skills the students get the chance to demonstrate depends on the linguistic opportunities they are given \cite{caines-buttery-2017}. Certainly we did find that student performance varied from lesson to lesson, so that the student might be B2 in one lesson for instance, and B1 or C1 in others. In future work, we wish to systematically examine the interplay between lesson structure, teaching feedback and student performance, because at present we can only observe that performance may vary from lesson to lesson.

The grammatical error correction performed on student turns in TSCC enables subsequent analysis of error types. We align each student turn with its corrected version, and then type the differences found according to the error taxonomy of \newcite{bryant-et-al-2017} and using the ERRANT program\footnote{\url{https://github.com/chrisjbryant/errant}}. We then count the number of instances of each error type and present them, following \newcite{bryant-et-al-2019}, as major edit types (`missing', `replacement' and `unnecessary' words) and grammatical error types which relate more to parts-of-speech and the written form. To show how TSCC compares to other error-annotated corpora, in Table~\ref{tbl:errors} we present equivalent error statistics for the FCE Corpus of English exam essays at B1 or B2 level \cite{yannakoudakis-et-al-2011} and the \textsc{CrowdED} Corpus of exam-like speech monologues by native and non-native speakers of English \cite{caines-et-al-2016}.

It is apparent in Table~\ref{tbl:errors} that in terms of the distribution of edits and errors the TSCC is more alike to another written corpus, the FCE, than it is to a speech corpus (\textsc{CrowdED}). For instance, there are far fewer `unnecessary' edit types in the TSCC than in \textsc{CrowdED}, with the majority being `replacement' edit types like the FCE. For the error types, there is a smaller catch-all `other' category for TSCC than \textsc{CrowdED}, along with many determiner, noun and preposition errors in common with FCE. There is a focus on the written form, with many orthography and spelling errors, but far fewer punctuation errors than the other corpora \textendash~a sign that chat interaction has almost no standard regarding punctuation.

In Figure~\ref{fig:sequences} we show where selected sequence types begin as points in the progress of each lesson (expressed as percentages) and which participant begins them, the teacher or student. Opening and closing sequences are where we might expect them at the beginning and end of lessons. The bulk of topic management occurs at the start of lessons and the bulk of eliciting and scaffolding occurs mid-lesson. Comparing the different CEFR levels, there are many fewer exercise and eliciting sequences for the C1 students compared to the B1 and B2 students; in contrast the C1 students do much more enquiry. In future work we aim to better analyse the scaffolding, repair and revision sequences in particular, to associate them with relevant preceding turns and understand what prompted the onset of these particular sequences.

\section{Conclusion}

We have described the Teacher-Student Chatroom Corpus, which we believe to be the first resource of its kind available for research use, potentially enabling both close discourse analysis and the eventual development of educational technology for practice in written English conversation. It currently contains 102 one-to-one lessons between two teachers and eight students of various ages and backgrounds, totalling 133K words, along with annotation for a range of linguistic and pedagogic features. We demonstrated how such annotation enables new insight into the language teaching process, and propose that in future the dataset can be used to inform dialogue system design, in a similar way to H\"{o}hn's work with the German-language deL1L2IM corpus \cite{hohn-2017}.

One possible outcome of this work is to develop an engaging chatbot which is able to perform a limited number of language teaching tasks based on pedagogical expertise and insights gained from the TSCC. The intention is not to replace human teachers, but the chatbot can for example lighten the load of running a lesson \textendash~taking the `easier' administrative tasks such as lesson opening and closing, or homework-setting \textendash~allowing the teacher to focus more on pedagogical aspects, or to multi-task across several lessons at once. This would be a kind of human-in-the-loop dialogue system or, from the teacher's perspective, assistive technology which can bridge between high quality but non-scalable one-to-one tutoring, and the current limitations of natural language processing technology. Such educational technology can bring the benefit of personalised tutoring, for instance reducing the anxiety of participating in group discussion \cite{griffin-roy-2019}, while also providing the implicit skill and sensitivity brought by experienced human teachers.

First though, we need to demonstrate that (a) such a CALL system would be a welcome innovation for learners and teachers, and that (b) chatroom lessons do benefit language learners. We have seen preliminary evidence for both, but it remains anecdotal and a matter for thorough investigation in future. Collecting more data of the type described here will allow us to more comprehensively cover different teaching styles, demographic groups and L1 backgrounds. At the moment any attempt to look at individual variation can only be that: our group sizes are not yet large enough to be representative. We also aim to better understand the teaching actions contained in our corpus, how feedback sequences relate to the preceding student turns, and how the student responds to this feedback both within the lesson and across lessons over time.

\section*{Acknowledgments}
This paper reports on research supported by Cambridge Assessment, University of Cambridge. Additional funding was provided by the Cambridge Language Sciences Research Incubator Fund and the Isaac Newton Trust. We thank Jane Walsh, Jane Durkin, Reka Fogarasi, Mark Brenchley, Mark Cresham, Kate Ellis, Tanya Hall, Carol Nightingale and Joy Rook for their support. We are grateful to the teachers, students and annotators without whose enthusiastic participation this corpus would not have been feasible.

\bibliography{bib}
\bibliographystyle{acl_natbib}

\end{document}